\documentclass[sn-mathphys-num]{sn-jnl}

\usepackage{graphicx}%
\usepackage{multirow}%
\usepackage{amsmath,amssymb,amsfonts}%
\usepackage{amsthm}%
\usepackage{mathrsfs}%
\usepackage[title]{appendix}%
\usepackage{xcolor}%
\usepackage{textcomp}%
\usepackage{manyfoot}%
\usepackage{booktabs}%
\usepackage{algorithm}%
\usepackage{algorithmicx}%
\usepackage{algpseudocode}%
\usepackage{listings}%

\usepackage{comment}
\usepackage{kbordermatrix}

\theoremstyle{thmstyleone}%
\newtheorem{theorem}{Theorem}
\newtheorem{proposition}[theorem]{Proposition}%

\theoremstyle{thmstyletwo}%
\newtheorem{example}{Example}%
\newtheorem{remark}{Remark}%

\theoremstyle{thmstylethree}%
\newtheorem{definition}{Definition}%

\begin{document}

\title[Article Title]{Multi-class Classifier based Failure Prediction with Artificial and Anonymous Training for Data Privacy}


\author*[1]{\fnm{Dibakar} \sur{Das}}\email{dibakard@acm.org}

\author[2]{\fnm{Vikram} \sur{Seshasai}}\email{vikrams@tejasnetworks.com}

\author[2]{\fnm{Vineet Sudhir} \sur{Bhat}}\email{vineetb@tejasnetworks.com}

\author[2]{\fnm{Pushkal} \sur{Juneja}}\email{pushkal@tejasnetworks.com}

\author[1]{\fnm{Jyotsna} \sur{Bapat}}\email{jbapat@iiitb.ac.in}

\author[1]{\fnm{Debabrata} \sur{Das}}\email{ddas@iiitb.ac.in}

\affil*[1]{\orgname{International Institute of Information Technology Bangalore}, \orgaddress{\street{26/C, Hosur Road, Electronic City Phase 1}, \city{Bangalore}, \postcode{560100}, \state{Karnataka}, \country{India}}}

\affil[2]{\orgname{Tejas Networks Ltd}, \orgaddress{\street{Plot No.25, JP Software Park, Electronic City Phase 1}, \city{Bangalore}, \postcode{560100}, \state{Karnataka}, \country{India}}}


\abstract{Failures in real-world deployed systems, e.g., optical routers in the internet, may lead to major functional disruption. Hence, the prevention of failures is very important. Prediction of such failures would be a further enhancement in this direction. However, the failure prediction mechanism is not designed for every deployment system. Subsequently, such a prediction mechanism becomes a necessity. In such a scenario, non-intrusive failure prediction based on available information, e.g., logs, etc., has to be designed. Conventionally, logs are mined to extract useful information (data sets) to apply artificial intelligence (AI)/machine learning (ML) techniques. However, extraction of data sets from raw logs can be an extremely time-consuming effort. Also, logs may not be made available due to privacy issues. This paper proposes a novel non-intrusive system failure prediction technique using available information from developers and minimal information from raw logs (rather than mining entire logs) but keeping the data entirely private with the data owners. A neural network-based multi-class classifier is developed for failure prediction, using an artificially generated anonymous data set, applying a combination of techniques, viz., genetic algorithm (steps), pattern repetition, etc., to train and test the network. The proposed mechanism completely decouples the data set used for the training process from the actual data which is kept private. Moreover, multi-criteria decision-making (MCDM) schemes are used to prioritize failures in meeting business requirements. Results show high accuracy in failure prediction under different parameter configurations. In a broader context, any classification problem, beyond failure prediction, can be performed using the proposed mechanism with an artificially generated data set without looking into the actual data as long as the input features can be translated to binary values (e.g. output from private binary classifiers) and can provide classification-as-a-service.
}




\maketitle

\section{Introduction}\label{sec1}
Real-world deployed systems fail due to various unforeseen reasons. Any such failure can lead to severe disruption of functionalities in the associated environment. For example, a failure of an optical router in the internet backbone can lead to major losses of revenue for the operators and businesses. Hence, it is essential to prevent system failures by predicting them before they happen so that mitigation or avoidance procedures can be actuated. Sometimes, failure prediction mechanisms are not always built into the design of the deployed systems. In such situations, a non-intrusive failure prediction mechanism becomes a necessity since neither major changes are possible in the deployed system nor recommended. This leads to a couple of challenges. Firstly, statistics of preceding events leading to the failures are not readily available. Secondly, even though raw logs are available, mining them to get the necessary information to apply conventional AI/ML-based techniques can be an extremely time-consuming exercise. Thirdly, system logs contain sensitive information about the product and hence may not be made available for mining due to privacy reasons.

System logs contain a wealth of information. Several research proposals have been put forward over decades to mine the logs and predict system failures \cite{cite_failure_predict_log_recent_survey}. Detecting anomalies in systems applying deep learning on logs has been proposed \cite{cite_anomaly_log_dl_survey}. A security vulnerability in systems applying log analysis has also been explored extensively \cite{cite_security_log_analysis_survey}. Predicting different types of failures in high-performance computing (HPC) systems has been widely deployed \cite{cite_hpc_failure_pred_survey}. Typically, these proposals use different kinds of recurrent neural networks (RNN), e.g., long short-term memory (LSTM) and techniques, such as, autoencoders, gaussian mixer models, support vector machines, etc. All these techniques are data-intensive techniques where large amounts of logs are mined to construct the data sets. These data sets are then used by the above techniques to predict failures. Unlike the proposed method in this paper, none of the above techniques work without the statistics of the actual data.

\begin{figure}[ht]
\centering
\includegraphics[width=\columnwidth]{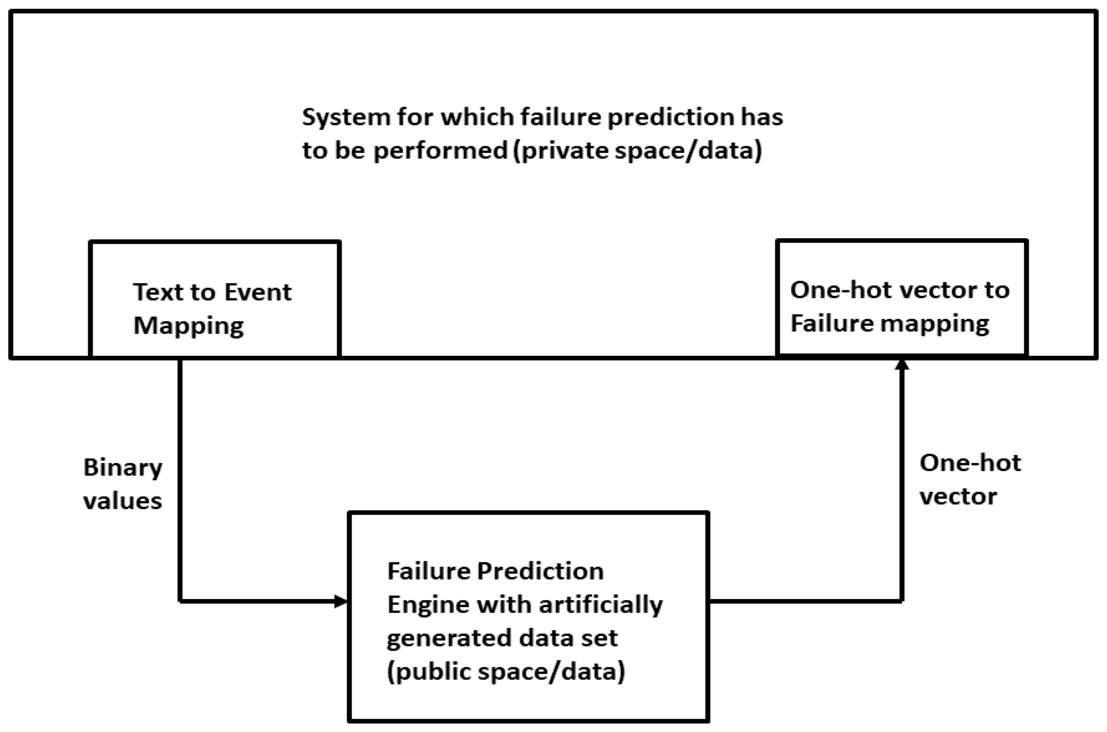}
\caption{Prediction with data privacy}
\label{fig_public_failure_pred_private_data_inkscape}
\end{figure}

This paper (preprint \cite{cite_preprint_anon_training_classifier}) presents a mechanism where key texts in the logs are designated as events and mapped onto binary values. This binary information is the input to the prediction engine. Each failure to be predicted as output is represented as a one-hot vector. This binary information is used to build a neural network (NN) based multi-class classifier for failure prediction. \emph{To train this classifier, an artificial data set is constructed using random sampling, pattern repetition, and steps from the genetic algorithm (GA) \cite{cite_genetic_algorithms_survey} without any knowledge about the actual data which is kept private.} The \emph{argmax} of the softmax output layer of the classifier is the predicted failure. To prioritize the failures based on business needs, an MCDM mechanism, viz., Analytical Hierarchical Process (AHP), is used to assign weights to failures. Both the weights (a vector) and the probabilities of the softmax layer (also a vector) are passed through a shape-preserving filter to reduce their variances. The \emph{argmax} of the product of corresponding elements of the two filtered vectors is the prioritized predicted failure. Results show that the proposed model predicts failures with high accuracy. \emph{Note that the mapping of the text events to the binary values can be kept private by the product/data owners. Also, the mapping from a one-hot vector to the actual failure can be kept private. Thus, the prediction mechanism works on completely anonymously mapped binary information and their sequences (in time), without looking into the actual data, and hence helps in data privacy (Fig. \ref{fig_public_failure_pred_private_data_inkscape}).}
For real-time failure prediction, logs are parsed through a time-based sliding window parser to look for events, they are mapped to binary values (by product/data owners) in the private domain and then passed to the public domain NN-based multi-class classifier for failure prediction. The predict one-hot vector is passed back to the private domain for handling. The key part is the public classifier completely working on an artificially generated data set. The current authors proposed two methods for failure prediction using directed acyclic graphs \cite{cite_self_dag_fail_pred} and data-augmented Bayesian networks \cite{cite_self_data_augment_bn_fail_pred} which do not scale up well with large number of events and failures. Also, NN provides faster inference compared to the two methods. The method proposed in this paper is highly scalable. To the best of the knowledge of the authors, none of the previous work in literature deals with this proposed novel multi-class classifier for system failure prediction and their prioritization using artificial data sets to maintain data privacy.

This model can be used in a collaborative setting.
To understand this aspect, lets consider a simple example. Consider three companies \emph{A}, \emph{B} and \emph{C} who want to cooperate on a classification problem but they do not want to share their model or data. Company \emph{A} has binary classifier which clearly recognizes a leopard and provide a binary output. Similarly, company \emph{B} has a binary classifier which recognizes a jungle. Also, company \emph{C} classifies a city. Suppose, they want to collaborate to classify whether the leopard is in the city or jungle. In such a case, the proposed classifier can do the job since it classifies sequences of binary inputs. Each of these inputs to our classifier can come as outputs of the private binary classifiers of \emph{A}, \emph{B} and \emph{C}. For example, if the outputs from \emph{A}, \emph{B} and \emph{\emph{C}} is 1, 1 and 0 respectively then the public classifier can be trained with these as inputs to classify a leopard in the jungle with output as a one hot vector. Similarly, if the outputs from \emph{A}, \emph{B} and C is 1, 0 and 1 respectively then the public classifier can be trained with these as inputs to classify a leopard in the city with output having a different one hot vector. During inference, the first case is normal but for the second case the system could raise an alarm.
If there are thousands of such binary scenarios/sequences which are typical in a large computer network the proposed approach can be very valuable and can help predicting failures, for instance, in heterogeneous equipment from different vendors who may not like to share their machine learning models or data. But, they may definitely say that they see certain (anomalous) events coming and raise flags which when corroborated with others can lead prediction of different network failures.

There are several advantages of this proposed model.
\begin{itemize}
\item No data sets from actual data need to be constructed for training the multi-class classifier. Training data is generated artificially by applying steps of GA, repetition, and random sampling. Spending time on mining the logs can be avoided.
\item Minimal input from developers, such as, text to event map, sequence of events (only one for each failure), and failure priority are needed. A sequence of events is represented by binary values. Only one sequence is required per failure.
\item It is a non-intrusive approach and does not make any changes to the deployed system.
\item Data privacy is ensured since the model does not look into the actual data.
\item This mechanism can be provided as a general service, independent of the actual product/data, as long as input features are mapped to binary values and binary one-hot vectors as outputs. For example, the output of binary classifiers in the private domain can be input to this proposed classifier.
\item In a broader context, any classification problem translated to binary inputs and binary one-hot vectors as outputs can be performed using the proposed architecture with an artificially generated data set, without looking into actual data and providing classification-as-a-service.
\item Input binary sequences can be reused and mapped to different data (in the private domain). As long as the sequences do not change, no new training is necessary.
\end{itemize}

This paper is organized as follows. Section \ref{section_literature_survey} contains a survey of recent works related to this proposal. The overall architecture of the failure prediction mechanism is described in section \ref{section_architecture}. The system model of the failure prediction engine is described in section \ref{section_system_model}. Results obtained from the failure prediction mechanism are discussed in section \ref{section_results}. Section \ref{section_conclusion} concludes this paper along with some future extensions.

\section{Literature survey}\label{section_literature_survey}
For high-performance computing (HPC), \cite{cite_hpc_dl_rnn_lstm_failure_predict_log} presents a long short-term memory (LSTM) based recurrent neural network (RNN) making use of log files to predict lead time to failures. An LSTM-based solution for mission-critical information systems analyzing logs has been presented in \cite{cite_it_system_failure_pred_lstm_rnn}. \cite{cite_failure_predict_logs_mlp_radial_basis_linear_kernels} presents a mechanism to predict failure sequences in logs in the context of telemetry and automobile sectors using multilayer perceptron, radial basis, and linear kernels. To predict events (leading to failure) in the system, analyzing multiple independent sources of times series data using different ML methods has been investigated in \cite{cite_time_series_multi_system_fault_mgmt}. Several proposals have been put forward to predict vulnerability and security-related events in systems. A detailed survey of this topic has been summarized in \cite{cite_security_log_analysis_survey}. Run time anomalies in applications using logs and ML-based techniques have been proposed in \cite{cite_application_anomalies_ml}. Failure prediction in network core routers analyzing logs, building a data set, and then applying support vector machines (SVM) has been proposed in \cite{cite_failure_predict_log_svm_router}. A multimodal anomaly detection applying unsupervised learning using a microphone, thermal camera, and logs in data center storage has been proposed in \cite{cite_anomaly_unsuprevised_multimodal_dc}. A lightweight training-free online error prediction for cloud storage applying tensor decomposition to analyze storage error-event logs has been proposed in \cite{cite_training_free_tensor_decompose_logs_storage}. A multi-layer bidirectional LSTM-based method for task failure prediction in a cloud computing environment using log files has been proposed in \cite{cite_task_failure_cloud_dl}.
\cite{cite_nlp_log_parsing} presents log parsing techniques using natural language processing for anomaly detection in aeronautical systems and public big data clusters (e.g., HDFS).
Applying non-parametric Chi-Square test and parametric Gaussian Mixture Model approaches, \cite{cite_mobile_outage_chi_square_guassian_mixture} proposes a mobile network outage prediction with logs.
An ML mechanism to predict job and task failures in multiple large-scale production systems from system logs has been described in \cite{cite_multiple_large_prod_ml_log}.
A decentralized online clustering algorithm based anomaly detection from resource usage logs of supercomputer clusters has been explored in \cite{cite_supercomputer_node_fail_pred_dist_cluster_algo}. Disk failure prediction in data centers using different ML techniques such as online random forests \cite{cite_dc_disk_fail_pred_online_random_forest}, auto encoders \cite{cite_improve_ssd_fail_pred_autoencoder} has also been proposed.  A predictive learning mechanism to detect latent error and fault localization in micro-service applications has been explored in \cite{cite_fault_pred_micro_services_pred_sys_logs}. Rare failure predictions is aircrafts using auto-encoder and bidirectional gated RNN mining failure logs have been explored in \cite{cite_aircraft_rare_failure_autoencoders_rnn}. An FP-Growth algorithm along with an adaptive sliding window division method to mine patterns in logs to predict failures has been proposed in \cite{cite_pattern_mining_failure_predict_large_cluster_logs}.
A recent survey of failure prediction based on log analysis is presented in \cite{cite_failure_predict_log_recent_survey}. Recent trends in anomaly detection using logs and applying deep learning have been surveyed in \cite{cite_anomaly_log_dl_survey}. A detailed survey of failure prediction in HPC (from logs) has been presented in \cite{cite_hpc_failure_pred_survey}. \textcolor{black}{Similarly intended approaches have also been applied to the field of cancelable biometrics using various non-invertible transformations \cite{cite_bioconvolving_signature_recognize}\cite{cite_bioconvolving_heuristic_selection}. A comprehensive survey of cancelable biometrics studying large number of techniques has been presented in \cite{cite_cancelable_biometrics_survey}.}

From the above survey of recent works in the field of failure prediction using various AI/ML techniques, a few points are evident. Firstly, all the mechanisms require large amounts of logs to be mined for building data sets to apply conventional AI/ML techniques. This approach involves significant effort to mine the logs. Secondly, logs contain very sensitive information and hence, may not be readily available under all circumstances. Thirdly, all the surveyed models use actual data. The novel approach proposed in this paper avoids the complete mining of logs and uses the information already available with the product developers and at the same time keeps the product data private to their owners. The entire training of the NN-based multi-class classifier for predicting failures is based on artificially generated data sets without looking into actual data. To the best of the knowledge of the authors, none of the prior research in the literature addresses the features of the proposed technique.
\section{Failure prediction architecture}\label{section_architecture}
The proposed mechanism can be used when failure prediction is not built into the initial design of the deployed system. This leads to a couple of challenges. Firstly, statistics of events and failures are not readily available. Secondly, systems are already deployed in the field and hence no changes can be made to incorporate a new failure prediction mechanism. Thirdly, detailed logs may not be available due to data privacy. In such as scenario, a non-intrusive failure prediction mechanism is the favoured approach making use of existing information. To incorporate a non-intrusive mechanism a few approaches are possible. Most systems have some form of logging mechanism to capture key events in the form of texts. Mining all the historical logs to extract useful data sets and then applying conventional AI/ML approaches can be one option to build a failure prediction engine (as discussed in section \ref{section_literature_survey}). However, extracting relevant information from raw logs can be an extremely time-consuming effort. Another approach can be to use only key information from logs with the help of the developers (rather than mining entire logs) and then build a failure prediction mechanism.  This approach offers a quicker as well as a non-intrusive solution. Hence, this approach is considered in this paper for deployed systems. Information provided by the developers includes text-to-event mapping (this can be kept private by the data owner), sequence of events leading to failures, and priority of each failure (pairwise relative importance of failures).
The current set of authors proposed two such non-intrusive failure prediction mechanisms \cite{cite_self_dag_fail_pred}\cite{cite_self_data_augment_bn_fail_pred}.
The architecture proposed in those works is further simplified, removing the need for a serializer as explained below.

\begin{figure}[ht]
\centering
\includegraphics[width=\columnwidth]{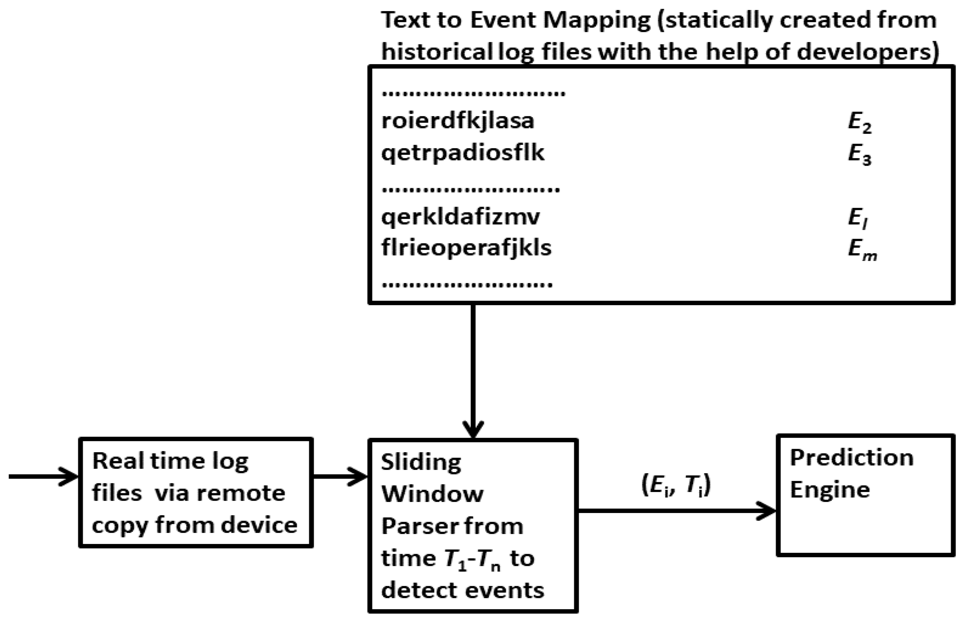}
\caption{Non-intrusive failure prediction architecture}
\label{fig_parser_arch_inkscape}
\end{figure}

\begin{figure}[ht]
\centering
\includegraphics[width=\columnwidth]{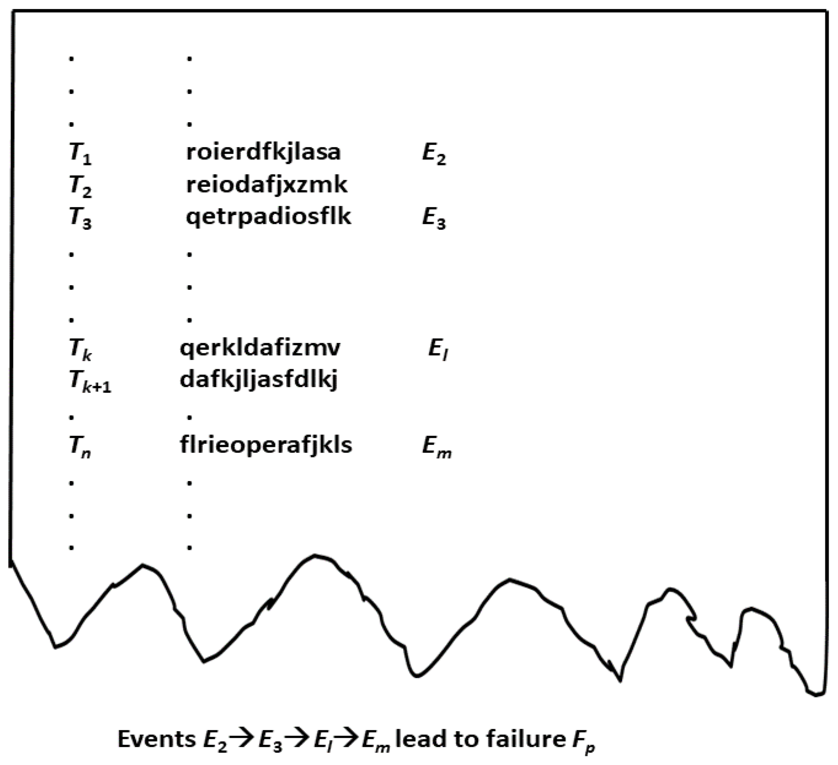}
\caption{Template device log file}
\label{fig_log_file_template_inkscape}
\end{figure}

The modified failure prediction architecture (Fig. \ref{fig_parser_arch_inkscape}) is discussed briefly. A device log typically consists of two columns, time and the corresponding texts, as shown in Fig. \ref{fig_log_file_template_inkscape}. Some of the key texts are designated as events with the help of the developers and added as a third column. 
A text-to-event mapping table (in Fig. \ref{fig_parser_arch_inkscape}) is constructed from this information. Each (suspected) failure can then be described as a sequence of these events. For predicting failures, real-time logs (either partial or whole) are read from the concerned system at regular intervals to a remote server. Using the text-to-event mapping table, the real-time logs are parsed using a time-based sliding window parser to look for events. This parser outputs the event and the corresponding time as a tuple to the prediction engine to predict failures. In previous research works \cite{cite_self_dag_fail_pred}\cite{cite_self_data_augment_bn_fail_pred} a serializer was used to order the events according to their times of occurrence from the tuples, which has been removed in this proposal. \emph{Note that all the above steps can be kept private by the product/data owners (Fig. \ref{fig_public_failure_pred_private_data_inkscape}). The classifier works on the binary inputs mapped to the tuples as will be explained subsequently.}
\section{System model of the multi-class classifier-based failure prediction}\label{section_system_model}
\begin{figure}[ht]
\centering
\includegraphics[width=\columnwidth]{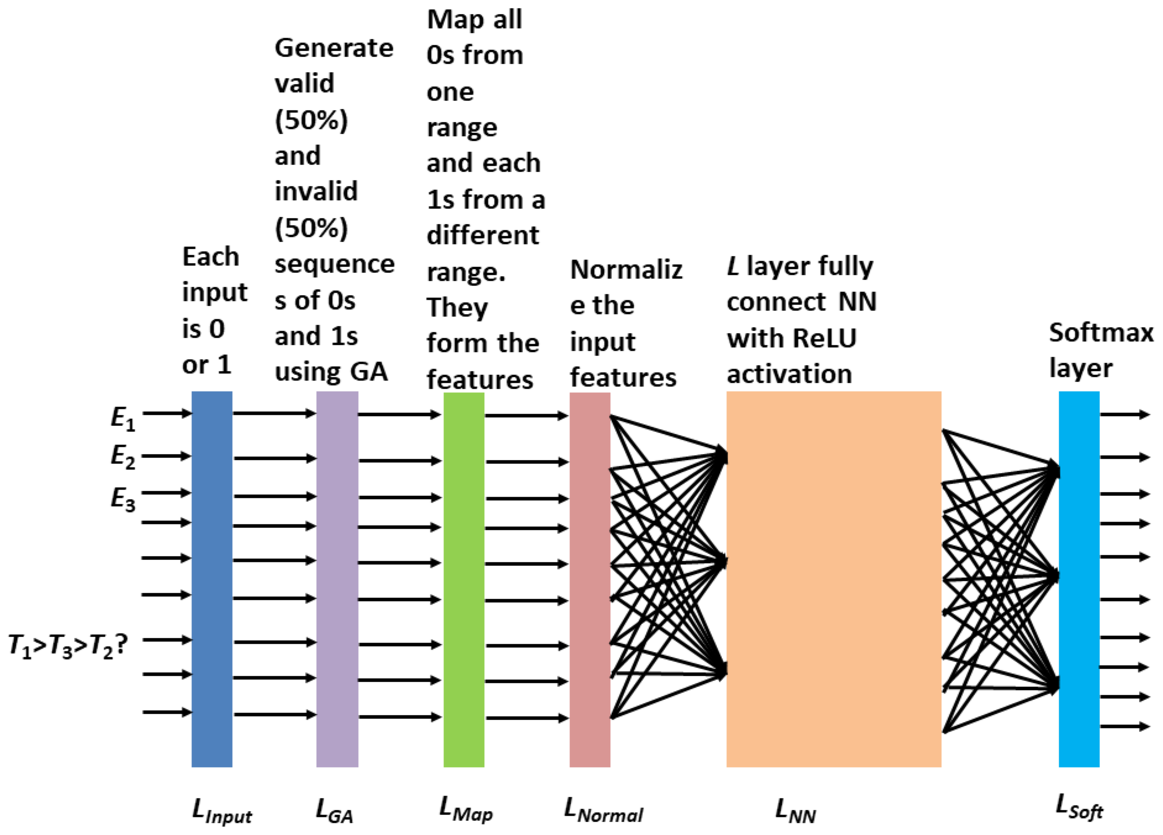}
\caption{Multi-class classifier-based failure prediction model using GA and NN}
\label{fig_predict_model}
\end{figure}

\begin{figure}[ht]
\centering
\includegraphics[width=\columnwidth]{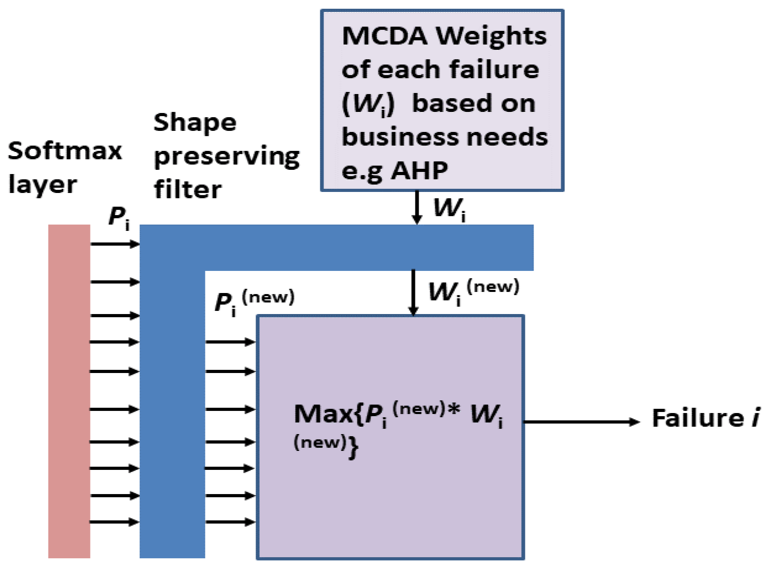}
\caption{Failure prioritization model using MCMD}
\label{fig_priority_model}
\end{figure}

The multi-class classifier-based failure prediction and prioritization models are shown in Figs. \ref{fig_predict_model} and \ref{fig_priority_model}. 
Each step of the model is explained in the sub-sections below.
\subsection{Input features}
Let's consider there are $F_{max}$ failures possible in the system. Each of these failures happens due to a sequence of events from a set of $E_{max}$ events. Each event has an associated time thus forming a $(E_i, T_i)$ tuple where event $E_i$ occurs at time $T_i$. For example, lets consider a failure $F_1$ which occurs when the following sequence of events occur, $E_2 \rightarrow E_5 \rightarrow E_8 \rightarrow E_{11}$ respectively at times $T_2$, $T_5$, $T_8$ and $T_{11}$. This implies $T_{11} > T_8 > T_5 > T_2$. There are also one-off events that may not have any time dependencies on any others. Each event is mapped to 1 when it occurs else it is 0.

Input features for training the classifier are the set of events, their sequences of occurrence in time leading to corresponding failures and one-off events. To incorporate the time dependencies of the events leading to the failures, their relationship is also added as an input feature. For example, failure $F_1$ mentioned above, the input features are the events $E_2$, $E_5$, $E_8$ and $E_{11}$, and their timing relationship, denoted as $(T_{11} > T_8 > T_5 > T_2)?$,  is satisfied or not which is again mapped to a binary value of 1 or 0 respectively. By default, each of times $T_2$, $T_5$, $T_8$ and $T_{11}$ is set to a value less than 0 (e.g., $T_{11} = T_8 = T_5 = T_2 = -1$), then the condition $(T_{11} > T_8 > T_5 > T_2)?$ is not satisfied, and hence the value is 0. Thus, $F_1$ occurs when $E_2 = 1$, $E_5 = 1$, $E_8 = 1$ and $E_{11} = 1$ and $(T_{11} > T_8 > T_5 > T_2)? = 1$ (i.e., the timing relation is also satisfied). For one-off events, the time of occurrence is ignored. In this way, there are a total of $E_{max}$ binary input features (0 or 1) which comprise of $E^{(rel)}_{max}$ related events corresponding $E^{(time)}_{max}$ timing relation events and $E^{(one)}_{max}$ one-off events. Thus,
\begin{equation}
E_{max} = E^{(rel)}_{max} + E^{(time)}_{max} + E^{(one)}_{max}
\end{equation}
These set of features of binary values form the input to $L_{Input}$ layer in Fig. \ref{fig_predict_model}. All the operations before the generation of binary values are kept private with the product/data owners (Fig. \ref{fig_public_failure_pred_private_data_inkscape}). \emph{Only information that is made public are the sequences of 1s and 0s (one sequence per failure). In the public domain, nothing can be inferred from what these binary values actually map to in the private domain. These binary sequences are enough to artificially and anonymously train the public NN-based multi-class classifier as will be explained subsequently.}

The binary representation of events as 1s and 0s is general enough to represent different kinds of data. Events such as incorrect system configurations and one-off events can be easily defined with binary values. For periodic failures, a separate event can be defined for each iteration. For events based on variation of certain parameters, going above or below certain thresholds, different events may be set when they cross a lower or an upper limit. The satisfiability of the timing relationship can also be represented as binary values (explained above). More generally, any feature value (e.g., temperature, pixel values, etc.) can be put into different buckets based on their variations in the private domain. For each bucket, a binary input can be defined in the public classifier. If the feature value belongs to a particular bucket then the corresponding bit is set to 1 else it is 0. The number of buckets and their ranges of values can vary based on the characteristics of the features. Binary sequences can also mean a state and the corresponding out one-hot vector of the NN can be the action to be taken. These binary inputs can in turn be the output of private binary classifiers or private multi-class classifiers with one-hot vectors as output. Also, having 1s and 0s helps in artificial data set generation as explained below.
\subsection{Artificial data set generation}\label{section_data_set_generate}
The statistics of the occurrence of failures and their corresponding events (features) are not known since no data set exists. In such a scenario, the only way is to create an artificial data set to train and test the NN-based multi-class classifier using the binary input generated above. For this purpose, the following three steps are applied.
\begin{enumerate}
\item Generation of data set to train and test the classifier applying repetition and steps from GA
\item Mapping each input binary feature to different values to create diversity in the data set
\item Normalization of the mapped data set
\end{enumerate}
These three steps form the layers $L_{GA}$, $L_{Map}$ and $L_{Normal}$ respectively in Fig. \ref{fig_predict_model}.
\subsubsection{Generation of training and test sets}\label{section_ga}
For $E_{max}$ features there are $2^{E_{max}}$ possible sequences of 1s and 0s. Out of these, only $F_{max}$ sequences are designated as failures which are far less than $2^{E_{max}}$. These $F_{max}$ sequences have to be predicted as valid failures by the classifier and the rest $2^{E_{max}} - F_{max}$ are to be classified as invalid failures. If the total input data set $S_{input}$ with its size denoted as $|S_{input}|$, it is divided into two sets of size $\frac{|S_{input}|}{2}$, one for valid failures and the other for invalid ones. For $\frac{|S_{input}|}{2}$ valid failures, each failure is repeated $\frac{|S_{input}|/2}{F_{max}}$ times. For $\frac{|S_{input}|}{2}$ invalid failures, the following steps from GA are applied, namely, selection, crossover, and mutation. For the \emph{selection} process, two sequences of input features from the valid failures are chosen at random. For \emph{crossover}, from the first failure the upper $\frac{E_{max}}{2}$ events are taken and from the second the lower $\frac{E_{max}}{2}$ are taken and concatenated together to form a new sequence. Then, for \emph{mutation}, some of the bits in the concatenated sequence are randomly selected and toggled. Since, there is no cost function to be optimized by the GA, iterations over multiple generations are not performed. Only the above three steps from GA are necessary for the current requirement. In this way, $\frac{|S_{input}|}{2}$ invalid failure sequences are generated. \emph{These steps from GA help in picking up "near by" sequences (which are more likely to occur) from the set of valid failures to train the classifier as invalid failures, rather than incorporating unlikely "far off" sequences from the much bigger $2^{E_{max}} - F_{max}$ set}.

To explain the application of GA, let's consider the following event to failure map in (\ref{eqn_event_failure_matrix_example}). For example, failure $F_3$ occurs when events occur in sequence  $E_1 \rightarrow  E_3 \rightarrow E_4 \rightarrow E_5$.
\begin{equation}
X = \kbordermatrix{
        & E_1 & E_2 & E_3 & E_4 & E_5 & E_6\\
    F_1 & 1 & 1 & 0 & 0 & 1 & 1\\
    F_2 & 0 & 1 & 0 & 1 & 1 & 0\\
    F_3 & 1 & 0 & 1 & 1 & 1 & 0\\
    F_4 & 1 & 1 & 0 & 0 & 1 & 1\\
    F_5 & 1 & 1 & 1 & 0 & 1 & 1\\
    F_6 & 1 & 0 & 1 & 0 & 1 & 0
  }
  \label{eqn_event_failure_matrix_example}
\end{equation}
For the selection step, two row vectors from $X$ in (\ref{eqn_event_failure_matrix_example}) are chosen at random. For example, let's assume $F_2$ and $F_5$ are the selected vectors. For crossover step, $E_1$, $E_2$, $E_3$ are taken from $F_2$ and $E_4$, $E_5$, $E_6$ are obtained from $F_5$, and a new vector is formed as shown in (\ref{eqn_crossover_vector_example}), which is obviously not a valid failure.
\begin{equation}
\kbordermatrix
{\\
& 0 & 1 & 0 & 0 & 1 & 1
}\label{eqn_crossover_vector_example}
\end{equation}
For the mutation step, the random number of bits in (\ref{eqn_crossover_vector_example}) are toggled (in this case $E_5$ is flipped from 1 to 0) as shown in (\ref{eqn_mutation_vector_example}).
\begin{equation}
\kbordermatrix
{\\
& 0 & 1 & 0 & 0 & 0 & 1
}\label{eqn_mutation_vector_example}
\end{equation}
This vector is added to the invalid failure set. If the vector in (\ref{eqn_mutation_vector_example}) ends up being a valid failure in (\ref{eqn_event_failure_matrix_example}), it is discarded and a new attempt is made following the above three GA steps. In this way, $\frac{|S_{input}|}{2}$ invalid failures are generated.
\subsubsection{Mapping the input features of the generated data set}\label{section_map}
\begin{figure}[ht]
\centering
\includegraphics[width=\columnwidth]{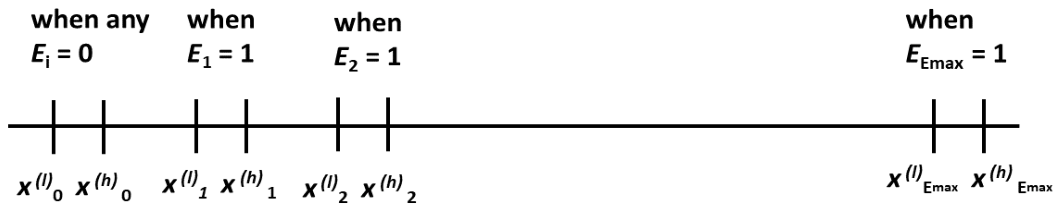}
\caption{Mapping an event to a random value}
\label{fig_event_to_value_map_inkscape}
\end{figure}

Each feature (or event) only has two values 0 or 1. These values are not very useful for training a NN-based classifier. Hence, each of the $E_{max}$ features is mapped to a positive uniform random value within a lower and an upper bound along a one-dimensional axis. All 0 inputs (i.e., $E_i = 0 $, $i=1,2,..,E_{max}$) are mapped to a single range $x^{(l)}_0$ to $x^{(h)}_0$ and each of the 1s (i.e., $E_i = 1 $, $i=1,2,..,E_{max}$) is mapped to separate ranges from $x^{(l)}_i$ to $x^{(h)}_i$ (Fig. \ref{fig_event_to_value_map_inkscape}). For example, if the failure vector after performing the GA steps is [1 0 0 1 0 1] then all the 0s are mapped to a uniform random value between 1.5 and 2.5. The first 1 is mapped to a value in the 5-10 range, the second 1 is mapped to 100-110 and the third 1 is mapped to 1000-1100, and so on. This process provides diversity to each input feature which helps in improved training of the multi-class classifier.
\subsubsection{Normalization}\label{section_norm}
Each feature (event) $x_i$ in the input vector after being mapped to a random value, as described above, is normalized by applying the min-max normalization in (\ref{eqn_min_max}).
\begin{equation}\label{eqn_min_max}
x^{'}_{i} = \frac{x_i- x_{min}}{x_{max} - x_{min}}
\end{equation}
where $x_{min}$ and $x_{max}$ are respectively the minimum and maximum values of the input feature vector.

The output labels $\hat{Y}$ of the multi-class classifier is a collection of $F_{max}$ one-hot row vectors with 1 set at the corresponding failure number position. The output label for all the generated invalid failures (using the GA steps) is a new one-hot vector $F_*$ with ${(F_{max} + 1)}^{th}$ position set to 1. Thus, the final $Y$ has dimension $(F_{max} + 1) \times (F_{max} + 1)$ and $F_{max}$ increases by one. These labels are attached as the data set $S_{input}$ is artificially generated (section \ref{section_ga}) before the mapping (section \ref{section_map}) and normalization (section \ref{section_norm}) steps explained above. Once, the output labels are attached to the generated $S_{input}$ set, the steps of mapping and normalization are performed on the binary input vectors.

For example, the output corresponding to the input in (\ref{eqn_event_failure_matrix_example}) is shown in (\ref{eqn_failure_output_example}). For the invalid failures, the output is a different one-hot vector other than those in (\ref{eqn_failure_output_example}). The dimension of $Y$ will change to $(F_{max} + 1) \times (F_{max} + 1)$ (i.e. $7 \times 7$) as shown in (\ref{eqn_failure_output_modified_example}). The vector $F_*$ underlined in (\ref{eqn_failure_output_modified_example}) is the output label for all the generated invalid failures.
\begin{equation}
\hat{Y} = \kbordermatrix{\\
    F_1 & 1 & 0 & 0 & 0 & 0 & 0\\
    F_2 & 0 & 1 & 0 & 0 & 0 & 0\\
    F_3 & 0 & 0 & 1 & 0 & 0 & 0\\
    F_4 & 0 & 0 & 0 & 1 & 0 & 0\\
    F_5 & 0 & 0 & 0 & 0 & 1 & 0\\
    F_6 & 0 & 0 & 0 & 0 & 0 & 1
  }
  \label{eqn_failure_output_example}
\end{equation}

\begin{equation}
Y = \kbordermatrix{\\
    F_1 & 1 & 0 & 0 & 0 & 0 & 0 & 0\\
    F_2 & 0 & 1 & 0 & 0 & 0 & 0 & 0\\
    F_3 & 0 & 0 & 1 & 0 & 0 & 0 & 0\\
    F_4 & 0 & 0 & 0 & 1 & 0 & 0 & 0\\
    F_5 & 0 & 0 & 0 & 0 & 1 & 0 & 0\\
    F_6 & 0 & 0 & 0 & 0 & 0 & 1 & 0\\
    \underline{F_*} & \underline{0} & \underline{0} & \underline{0} & \underline{0} & \underline{0} & \underline{0} & \underline{1}
  }
  \label{eqn_failure_output_modified_example}
\end{equation}

After the above steps, this artificial data set of inputs and outputs are permuted randomly and divided into training and test sets to be used by the NN-based classifier shown as $L_{NN}$ in Fig. \ref{fig_predict_model}. \emph{Note that the one-hot vectors in the public domain can be mapped to a failure which can be kept private by the product/data owner (Fig. \ref{fig_public_failure_pred_private_data_inkscape}). The one-hot vector derived from the softmax prediction can be forwarded to all the data product/data owners who can interpret the classification result in their own way. The classifier in the public domain can provide a classification-as-a-service, just like a processor executes binary instructions completely agnostic of the application, as along as private data can be mapped to binary inputs and outputs. The classifier need not know what those 0s and 1s map to in the private domain. Also, each binary bit can be reused and mapped to different events in the private domain without a need to retrain the classifier. For example, the same input bit can used for temperature exceeding a threshold in one classification problem and reused for signal strength going below a threshold in another without the need of training the classifier again as long as the sequences of 0s and 1s remain the same. As many binary sequences from $2^{E_{max}}$ patterns as needed can be shared among the users of the classifier. The data with owners in the private domain can be multi-modal. The output of private binary classifiers on those data may work as input to the proposed public classifier as binary values. This shared mechanism can work for collaborative classification keeping data of each owner private.}
\subsection{Fully connected neural network}
The NN block ($L_{NN}$) is constructed using a fully connected neural network consisting of $L_{nn}$  layers (input + hidden) with $E_{max}$ inputs and outputs. The output is a softmax (\ref{eqn_softmax}) layer ($L_{soft}$) of $E_{max}$ inputs and $F_{max}$ outputs.
\begin{equation}\label{eqn_softmax}
softmax(\hat{y}_i^{(k)}) = \frac{e^{\hat{y}_i^{(k)}}}{\sum_{k=1}^{F_{max}}e^{\hat{y}_i^{(k)}}}
\end{equation}
where $\hat{y}_i^{(k)}$ is the predicted probability of $F_k$ corresponding to the $i^{th}$ input example of size $E_{max}$ to $L_{nn}$ layers.

The NN is trained for multiple epochs ($N_{epochs}$) to minimize the cross entropy error function (\ref{eqn_cross_entropy}) over the training set $S_{train}$ of size $|S_{train}|$.
\begin{equation}\label{eqn_cross_entropy}
L(Y,\hat{Y}) = - \sum_{i=1}^{|S_{train}|}\sum_{k=1}^{F_{max}} y_i^{(k)}log(\hat{y}_i^{(k)})
\end{equation}
where $\hat{Y}$ is the predicted distribution of $Y$, $\hat{y}_i^{(k)}$ is the predicted value of $y_i^{(k)}$ for the failure $F_k$ corresponding to the $i^{th}$ training example.
Rectified linear unit (ReLU) activation function (\ref{eqn_relu}) is used in all the neurons of $L_{NN}$ block.
\begin{equation}\label{eqn_relu}
z^{(l+1)}_j = max(0,z^{(l)}_j)
\end{equation}
for $j = 1,2, .., E_{max}$ and $l = 1,2,..,L_{nn}$ and $z^{(l+1)}_j$ is the output of the $j^{th}$ neuron in $l^{th}$ layer of $L_{NN}$ block.
\subsection{Failure prioritization}
Often, technical decisions do not agree with business requirements. Hence, a failure that is predicted with high probability by the multi-class classifier may not be the one of highest priorities from a business perspective. For example, lets consider two arbitrary failures $F_m$ and $F_n$ with softmax probabilities $p_{F_m}$ and $p_{F_n}$ respectively and $p_{F_m} > p_{F_n}$. However, from a business perspective, $F_n$ is more important and has to be given higher priority than $F_m$. For prioritizing the failures each of them has to be assigned a weight based on business needs. To assign these weights, human judgment has to be incorporated into the model. Multi-criteria decision-making (MCDM) techniques help in this direction. Various MCDM techniques are proposed in the literature \cite{cite_multi_criteria_decision_survey}.
Among them, this work assigns weights to each of the failures applying AHP (though other techniques can be applied which will be considered in future work).

One way to decide weights is to make an arbitrary assignment for each item (failures). However, when there are many items to compare amongst, deciding the weights arbitrarily may not be the best option and is also non-trivial. The pairwise comparison works better in such cases. AHP applies pairwise comparison to decide weights for each item incorporating human judgement to prioritize the failures.

To prioritize the failures, a $F_{max} \times F_{max}$ matrix $C_{pair}$ is constructed for $F_{max}$ failures. Based on the business needs, each pair of failures is compared. For example, if $F_{n}$ is twice as important as $F_m$ then $C_{pair}[n,m]$ is assigned a value of $\frac{2}{1}$. $C_{pair}[m,n]$ is assigned $\frac{1}{2}$ which is reciprocal of $C_{pair}[n,m]$. In this way, the entire $C_{pair}$ matrix is initialized. All $C_{pair}[m,m]$ entries are set to 1. The principal eigen vector corresponding to the highest eigen value (evaluated using the power method \cite{cite_eigen_power_method}), provides the weights $w_{F_k}, k = 1,2,..,F_{max}$, for each failure.

\subsection{Shape preserving filtering}\label{section_shape_preserve_filter}
$L_{Soft}$ produces a vector of probabilities where some of the values may be higher compared to others. Similarly, some of the weights provided by AHP may also have comparatively higher values than the rest. If the elementwise products of the weights and the corresponding probabilities of the failures are calculated, then either of the factors can skew their product to a high value in favour of one failure.
Hence, it is necessary to bring the probabilities and weights of all the failures to a level playing field, so that their elementwise products are not dominated by either of the factors. For this purpose, a shape-preserving filtering mechanism is applied as explained below. In the current context, shape preserving means the reduction of variance among the values keeping relative ordering among them unaffected. The importance of this filtering will be evident from the results in section \ref{section_results}.

An iterative algorithm is developed to reduce the variance as well as preserve the order of the values. The steps are explained as follows. 
\begin{enumerate}
\item Initialize set of values to be filtered, $\Lambda = \{\lambda_1, \lambda_2,.., \lambda_k\}$
\item Initialize the value of $\delta$ which is at least an order of magnitude smaller than the values in $\Lambda$.
\item Find out the maximum value of $\Lambda$, $\lambda_{max} = max(\Lambda)$
\item Calculate $\lambda^{(new)}_{max} = \lambda_{max} - \delta$
\item Evaluate  $\Lambda^{(new)} = \{\{{\Lambda \setminus \{\lambda_{max}\}}\} + \delta\} \cup \{\lambda^{(new)}_{max}\}$. Here, "+" operation means addition of $\delta$ to each element of $\{{\Lambda \setminus \{\lambda_{max}\}}\}$
\item  If there is a violation of the relative ordering of values then return $\Lambda$ else $\Lambda = \Lambda^{(new)}$ go to step 3.
\end{enumerate}
$\Lambda$ contains the filtered values.
This algorithm is applied to both the output softmax probability vector of the $L_{Soft}$ layer and the weights derived from AHP.
Values from filtered weight vector $w^{(f)}_{F_k}$  and filtered probability vector $p^{(f)}_{F_k}$ are multiplied elementwise. The  $argmax\{w^{(f)}_{F_k}*p^{(f)}_{F_k}\}$ is the predicted failure $F_k$ satisfying business requirements.
\subsection{Performance matrices for the NN-based multi-class classifier}
$S_{input}$ denotes the set of inputs to the classifier (generated in section \ref{section_data_set_generate}) with cardinality $|S_{input}|$. $S_{input}$ is split into two sets, one of them is $S_{train}$ (with cardinality $|S_{train}|$) for training the classifier, and the other $S_{test}$ (with cardinality $|S_{test}|$) is used as test set. Hence,
\begin{equation}
|S_{input}| = |S_{train}| + |S_{test}|
\end{equation}
Performance is evaluated by counting the failure prediction errors on $S_{test}$ at the softmax layer output.  If $N^{(error)}_{test}$ is the number of failure prediction errors then the percentage is defined as,
\begin{equation}
P_{error} = \frac{N^{(error)}_{test}}{|S_{test}|} \times 100\%
\end{equation}
The final prioritized output after applying AHP is considered in detail latter in section \ref{section_results}.
\begin{table}[ht]
  \caption{Key parameters of the failure prediction model}
  \centering
  \begin{tabular}{|p{1.5cm}|p{11.5cm}|}
  \hline
  Parameter & Description\\  [0.5ex]
  \hline
  $F_{max}$ & Maximum number of failures\\
  \hline
  $E_{max}$ & Maximum number of events\\
  \hline
  $N^{(E)}_{F_k}$ & Number of events in $k^{th}$ failure \\
  \hline
  $L_{hidden}$ & Number of hidden layers apart from input and output softmax layer, i.e., $L_{hidden} = L_{NN} - 1$ \\
  \hline
  $\alpha^{(low)}_{events}$ & Minimum number of events set to 1 in a failure, $N^{(E)}_{F_k} \geqslant \alpha^{(low)}_{events} \times E_{max}$ \\
  \hline
  $\alpha^{(high)}_{events}$ & Maximum number of events set to 1 in a failure, $N^{(E)}_{F_k} \leqslant \alpha^{(high)}_{events} \times E_{max}$\\
  \hline
  $N_{epochs}$ & Number of training epochs\\
  \hline
  $M_{batch}$ & Mini batch size for training\\
  \hline
  $D_{thres}$ & Decision threshold applied for the probabilities at softmax output to decide whether the failure prediction is valid or invalid \\
  \hline
  \end{tabular}
  \label{table_model_parameters}
\end{table}
\subsection{Performance evaluation methodology}
The model is evaluated under general settings. Each failure $F_k$ is uniform randomly generated with 0s and 1s, so that number of 1s (i.e., events) in a failure $N^{(E)}_{F_k}$ lies in the range, $(\alpha^{(low)}_{events} \times E_{max}) \leqslant N^{(E)}_{F_k} \leqslant (\alpha^{(high)}_{events} \times E_{max})$, where $0\% < \alpha^{(low)}_{events} < \alpha^{(high)}_{events} < 100\% $. These failure vectors are passed to $L_{Input}$ layer of the classifier in Fig. \ref{fig_predict_model}. The output of  $L_{Normal}$ generates the set $S_{input}$. The output is a one-hot vector for each failure. Training of classifier ($L_{NN}$ + $L_{Soft}$) is performed to minimize the loss function (\ref{eqn_cross_entropy}) for $N_{epochs}$ with set $S_{train}$. The performance is evaluated on $S_{test}$ to measure $P_{error}$ (averaged over several iterations). The softmax output vector of probabilities and the AHP-generated weight vectors are then passed to the failure prioritization block in Fig. \ref{fig_priority_model} to decide on the failure priority according to business needs.
\section{Results and Discussion}\label{section_results}
This section discusses the results of the performances of the multi-class classifier used for failure prediction and their prioritization under different parameter configurations. Implementation is done using \verb|keras/tensorflow| framework.

The functions of $L_{GA}$, $L_{Map}$ and $L_{Normal}$ layers (Fig. \ref{fig_predict_model}) are performed as explained above with examples in section \ref{section_data_set_generate}. Performances of  $L_{Input}$, $L_{NN}$ and $L_{Soft}$ layers, and the prioritization block (Fig. \ref{fig_priority_model}) are studied under different configurations.
\subsection{Impact of test set size}\label{section_test_set_size}
\begin{figure}[ht]
\centering
\includegraphics[width=\columnwidth]{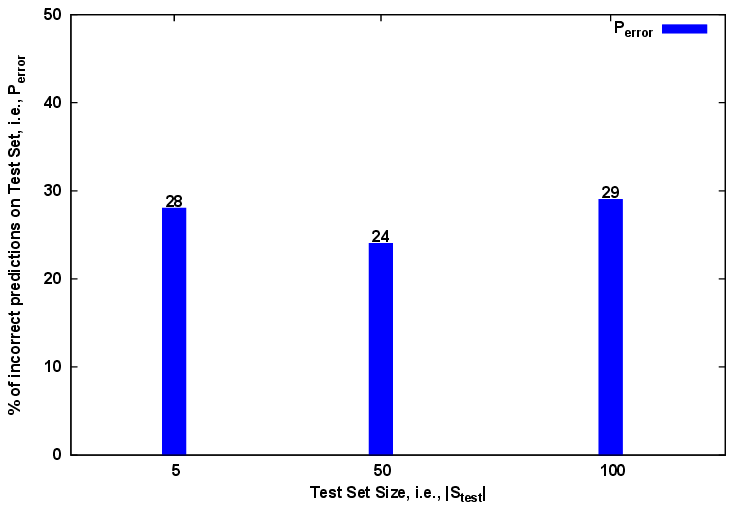}
\caption{Impact of increase in test set size (i.e., decrease in training set size)keeping the input set size constants}
\label{fig_increase_test_set_size_inkscape}
\end{figure}

In this section, the impact of increase in test set size $|S_{test}|$ (which means decrease in training set size $|S_{train}|$) is analyzed, keeping $|S_{input}|$ constant.
The configuration parameters are provided in Table \ref{table_test_set_size}.
\begin{table}[ht]
  \caption{Parameters of the failure prediction model for section \ref{section_test_set_size}}
  \centering
  \begin{tabular}{|p{2cm}|p{3cm}|}
  \hline
  Parameter & Values \\  [0.5ex]
  \hline
  $F_{max}$ & 50 \\
  \hline
  $E_{max}$ & 50 \\
  \hline
  $L_{hidden}$ & 5\\
  \hline
  $\alpha^{(low)}_{events}$ & 50\%\\
  \hline
  $\alpha^{(high)}_{events}$ & 80\%\\
  \hline
  $N_{epochs}$ & 40\\
  \hline
  $M_{batch}$ & 100 \\
  \hline
  $|S_{input}|$ & 500\\
  \hline
  $D_{thres}$ & 0.5\\
  \hline
  \end{tabular}
  \label{table_test_set_size}
\end{table}
For a small input set size ($|S_{input}|$)  of 500, the impact of the increase in test set size ($|S_{test}|$) is studied. In Fig. \ref{fig_increase_test_set_size_inkscape}, $|S_{test}|$ is shown along \emph{x}-axis and \emph{y}-axis depicts the percentage of incorrect predictions $P_{error}$ on $S_{test}$. It can be observed that for $|S_{test}|$ = 5 (i.e., $|S_{train}|$ = 500 - 5 = 495) $P_{error}$ is 28\%. For $|S_{test}|$ = 50 (i.e., $|S_{train}|$ = 500 - 50 = 450) $P_{error}$ = 24\%. If $|S_{test}|$ = 100 (i.e., $|S_{train}|$ = 500 - 100 = 400) then  $P_{error}$ = 29\%. Thus, configuration $|S_{test}|$ = 50 performs best. Results with $|S_{test}|$ = 5 underperform due to overfitting and those with $|S_{test}|$ = 100 also show degradation in performance due to less training data.
\subsection{Impact of training set size}
\begin{figure}[ht]
\centering
\includegraphics[width=\columnwidth]{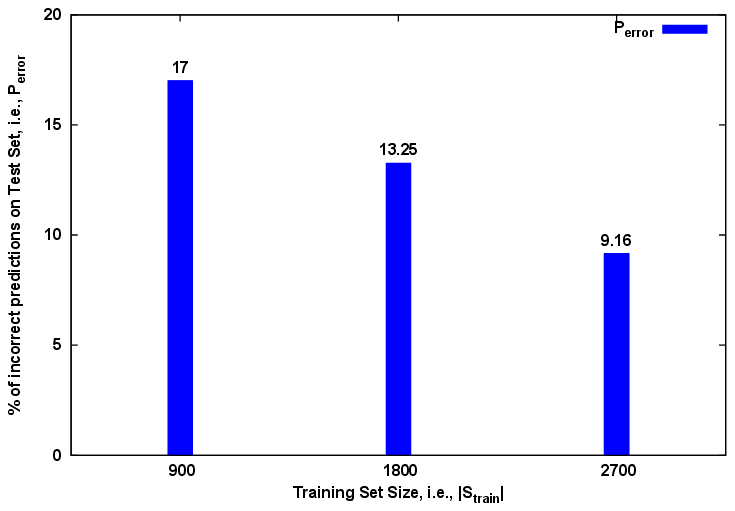}
\caption{Impact of increase in training set size}
\label{fig_increase_train_set_size_inkscape}
\end{figure}

In this section, $|S_{input}|$ is increased as 1000, 2000, and 3000, and then split into sets $S_{train}$ and $S_{test}$ in proportions of 90\% and 10\% respectively. Note that both $|S_{train}|$ and $|S_{test}|$ increase with larger $|S_{input}|$ although their proportions remain the same.
Increments in $|S_{train}|$ are shown along \emph{x}-axis and $P_{error}$ is depicted along \emph{y}-axis in Fig. \ref{fig_increase_train_set_size_inkscape}. The rest of the parameters remain the same as in Table \ref{table_test_set_size}. It can be observed that with larger artificially generated training data (90\% of $|S_{input}|$), the failure prediction on the larger test set (10\% of $|S_{input}|$) improves further.
\subsection{Impact of number of events per failure}
\begin{figure}[ht]
\centering
\includegraphics[width=\columnwidth]{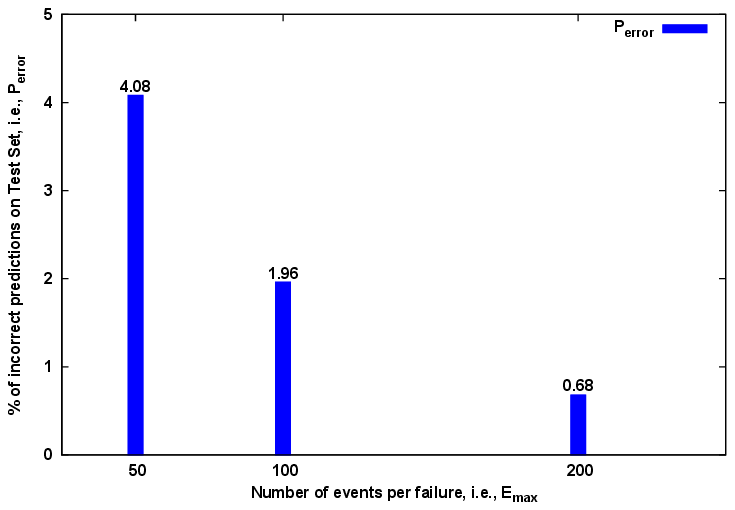}
\caption{Impact of increase in number events per failure}
\label{fig_increase_number_of_events_per_failure_inkscape}
\end{figure}

For $|S_{input}| = 5000$, $|S_{train}| = 4500$ and $|S_{test}| = 500$, maximum number of events per failure ($E_{max}$) is increased as 50, 100 and 200 along \emph{x}-axis in Fig. \ref{fig_increase_number_of_events_per_failure_inkscape} and \emph{y}-axis remains the same ($P_{error}$). Other parameters remain the same as in Table \ref{table_test_set_size}. Improvement in failure prediction is observed with a higher number of input features for the classifier.
\subsection{Impact of number of failures}
\begin{figure}[ht]
\centering
\includegraphics[width=\columnwidth]{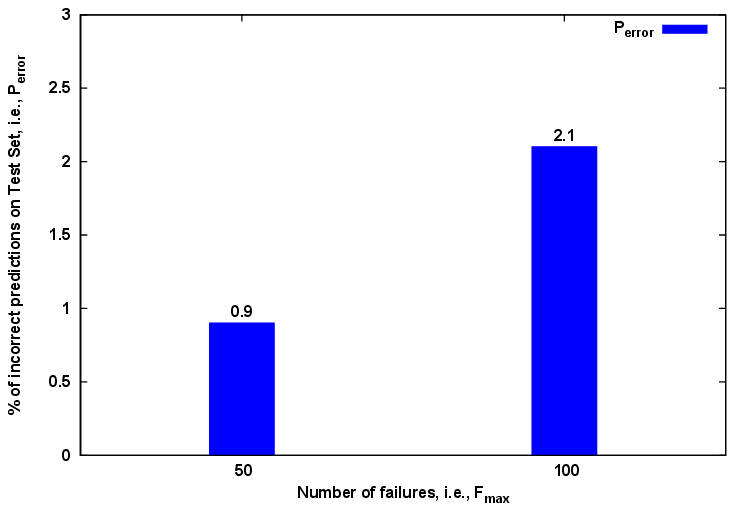}
\caption{Impact of increase in number of failures}
\label{fig_increase_number_of_failures_inkscapes}
\end{figure}

\begin{figure}[ht]
\centering
\includegraphics[width=\columnwidth]{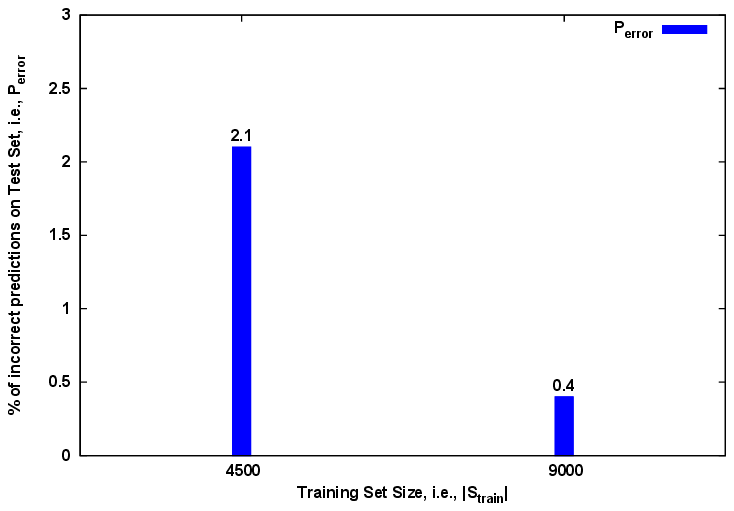}
\caption{Impact of increase in the number of failures with higher training data}
\label{fig_increase_number_of_failures_and_train_inkscape}
\end{figure}

In this subsection, the maximum of failures ($F_{max}$) is increased from 50 to 100 along \emph{x}-axis (Fig. \ref{fig_increase_number_of_failures_inkscapes}). The rest of the parameters remain the same as the previous configuration. It can be observed that such an increase in $F_{max}$ leads to degradation in the performance of the prediction engine (along \emph{y}-axis).
However, this degradation can be addressed by increasing $|S_{train}|$ from 4500 to 9000 for $F_{max} = 100$ (Fig. \ref{fig_increase_number_of_failures_and_train_inkscape}).
\subsection{Impact of increase in concentration of events in a failure}
\begin{figure}[ht]
\centering
\includegraphics[width=\columnwidth]{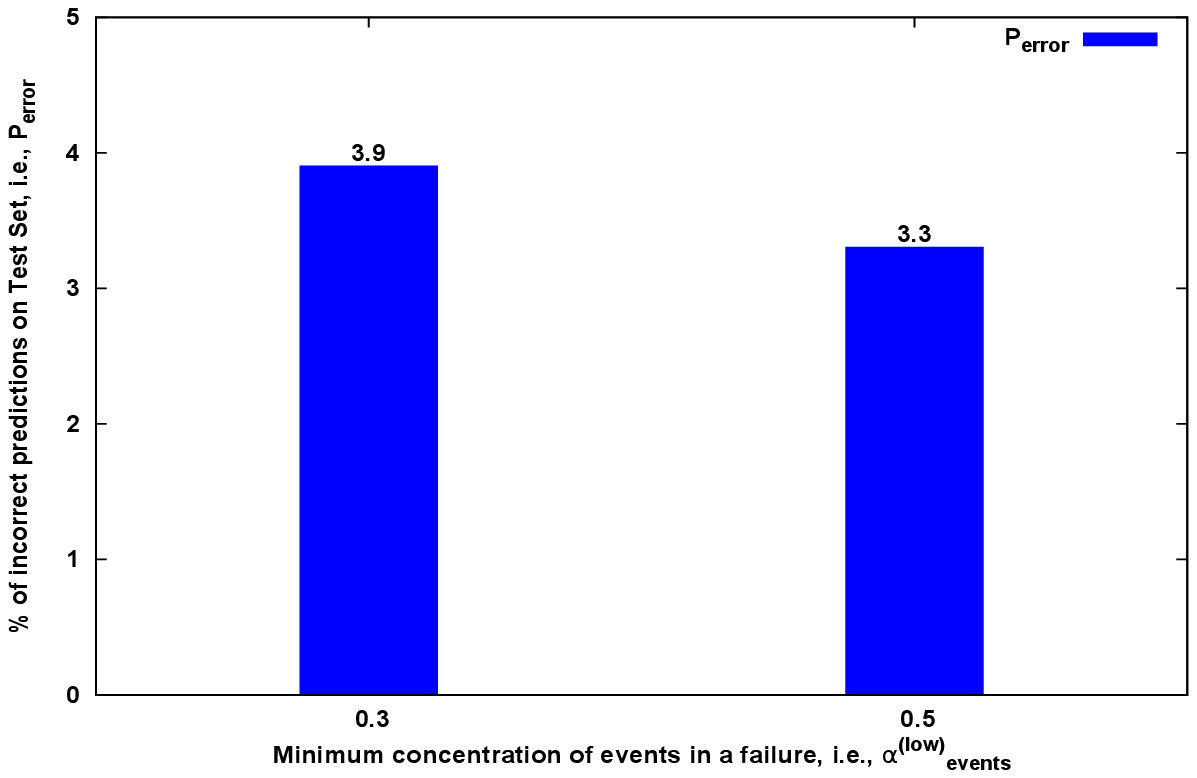}
\caption{Impact of increase in concentration of events in a failure}
\label{fig_increase_concentrate_events_inkscape}
\end{figure}

The concentration of events in a valid failure is increased by extending the interval between $\alpha^{(low)}_{events}$ and $\alpha^{(high)}_{events}$. The value of $\alpha^{(high)}_{events}$ is kept at 80\% but $\alpha^{(low)}_{events}$ is lowered to 30\% (depicted along \emph{x}-axis in Fig. \ref{fig_increase_concentrate_events_inkscape}). The rest of the parameters remain the same as before. If the concentration of the events is increased, this leads to marginally higher $P_{error}$ (along \emph{y}-axis), due to higher variance in the input feature set.
\subsection{Impact of increase in number of epochs}
\begin{figure}[ht]
\centering
\includegraphics[width=\columnwidth]{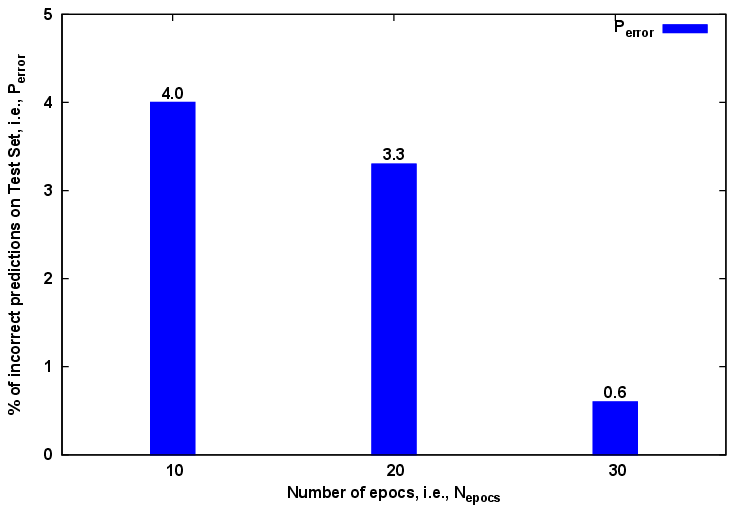}
\caption{Impact of increase in number of epochs}
\label{fig_increase_increase_epocs_inkscape}
\end{figure}

In Fig. \ref{fig_increase_increase_epocs_inkscape}, the training epochs ($N_{epochs}$) are changed as 10, 20, and 30 respectively along \emph{x}-axis and $P_{error}$ is plotted along \emph{y}-axis. It can be observed that a higher number of epochs leads to better training of the classifier leading to improved failure prediction.
\subsection{Impact of increase in mini-batch size}
\begin{figure}[ht]
\centering
\includegraphics[width=\columnwidth]{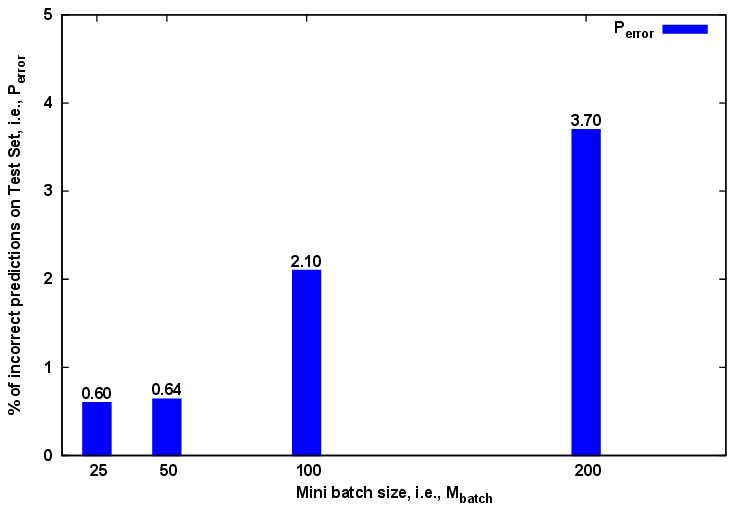}
\caption{Impact of increase in mini-batch size}
\label{fig_increase_mini_batch_size_inkscape}
\end{figure}

\begin{figure}[ht]
\centering
\includegraphics[width=\columnwidth]{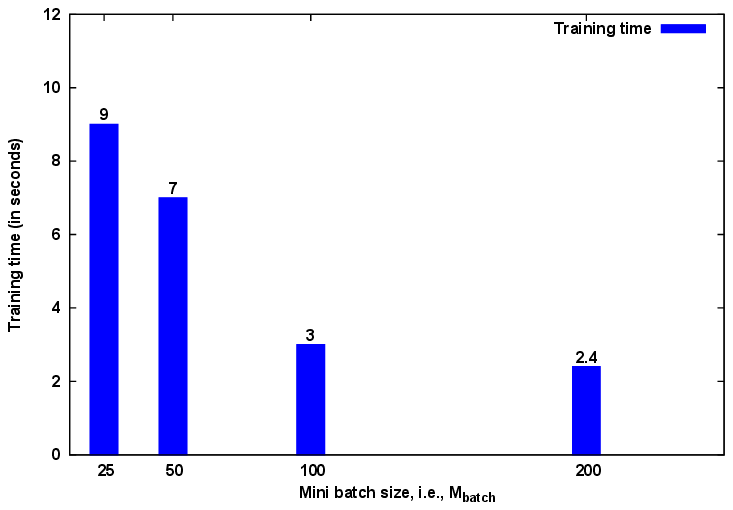}
\caption{Impact of increase in mini-batch size on training time}
\label{fig_increase_mini_batch_size_reduced_train_time_inkscape}
\end{figure}

Fig. \ref{fig_increase_mini_batch_size_inkscape} shows the impact of mini-batch size ($M_{batch}$) used by the gradient descent algorithm. The mini-batch size is changed along \emph{x}-axis and $P_{error}$ is shown along \emph{y}-axis.
With the increase in mini-batch size ($M_{batch}$), there is a degradation in performance with an increase in $P_{error}$. However, there is also a reduction in training time with an increase in $M_{batch}$ (Fig. \ref{fig_increase_mini_batch_size_reduced_train_time_inkscape}).
\subsection{Impact of decision threshold of softmax layer}\label{results_softmax_decision_thres}
\begin{figure}[ht]
\centering
\includegraphics[width=\columnwidth]{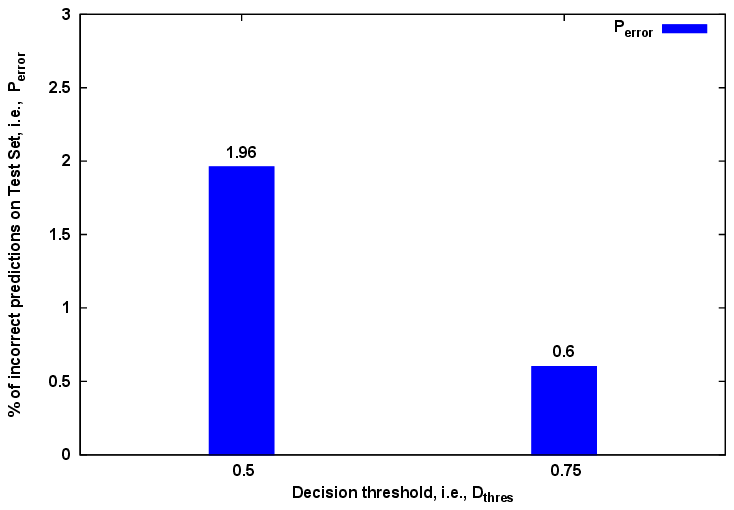}
\caption{Impact of softmax decision threshold}
\label{fig_increase_decision_threshold_inkscape}
\end{figure}

Fig. \ref{fig_increase_decision_threshold_inkscape} shows the impact of changing the decision threshold, i.e., $D_{thres}$, for softmax layer probabilities (along \emph{x}-axis) to decide whether the failure predicted is valid (at least one of the probabilities $\geqslant D_{thres}$) or invalid (all probabilities  $< D_{thres}$). It can be observed that $P_{error}$ is lower for $D_{thres}$ = 0.75 than when $D_{thres}$ = 0.50. This improvement in performance for $D_{thres}$ = 0.75 is because even for invalid failure prediction the softmax probabilities are quite high ($\geqslant 0.5)$ due to greater correlation among the event sequences of failures.
\subsection{Impact of number of hidden layers in the classifier}\label{section_results_hidden_layers}
\begin{figure}[ht]
\centering
\includegraphics[width=\columnwidth]{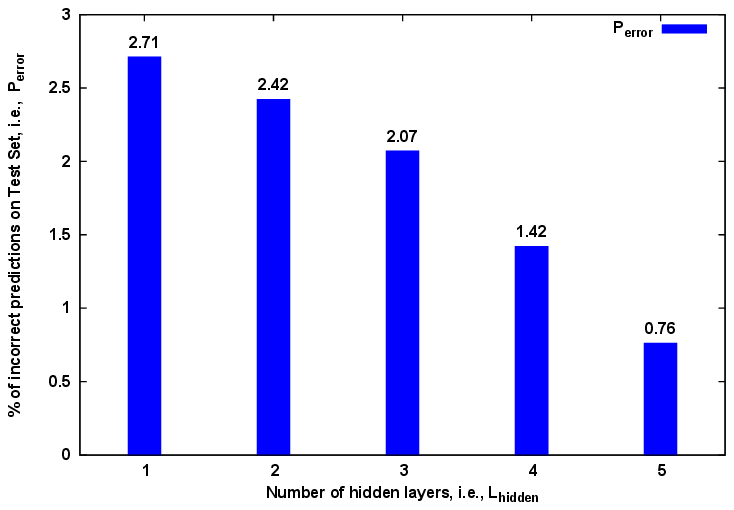}
\caption{Impact of increase in number of hidden layers}
\label{fig_increase_hidden_layers_inkscape}
\end{figure}

For large data sets with the parameters from Table \ref{table_results_hidden_layers}, the performance impact of adding hidden layers is shown in Fig. \ref{fig_increase_hidden_layers_inkscape}. Number of hidden layers ($L_{hidden}$) is shown along \emph{x}-axis and $P_{error}$ along \emph{y}-axis. It can be observed that with an increase in the number of hidden layers performance of the classifier improves.
\begin{table}[ht]
  \caption{Parameters for section \ref{section_results_hidden_layers}}
  \centering
  \begin{tabular}{|p{2cm}|p{3cm}|}
  \hline
  Parameter & Values \\  [0.5ex]
  \hline
  $F_{max}$ & 1000 \\
  \hline
  $E_{max}$ & 3000 \\
  \hline
  $L_{hidden}$ & 1-5\\
  \hline
  $\alpha^{(low)}_{events}$ & 50\%\\
  \hline
  $\alpha^{(high)}_{events}$ & 80\%\\
  \hline
  $N_{epochs}$ & 10\\
  \hline
  $M_{batch}$ & 1000 \\
  \hline
  $|S_{input}|$ & 100,000\\
  \hline
  $D_{thres}$ & 0.5\\
  \hline
  \end{tabular}
  \label{table_results_hidden_layers}
\end{table}
\subsection{Impact of learning rate}\label{results_learning_rate}
\begin{figure}[ht]
\centering
\includegraphics[width=\columnwidth]{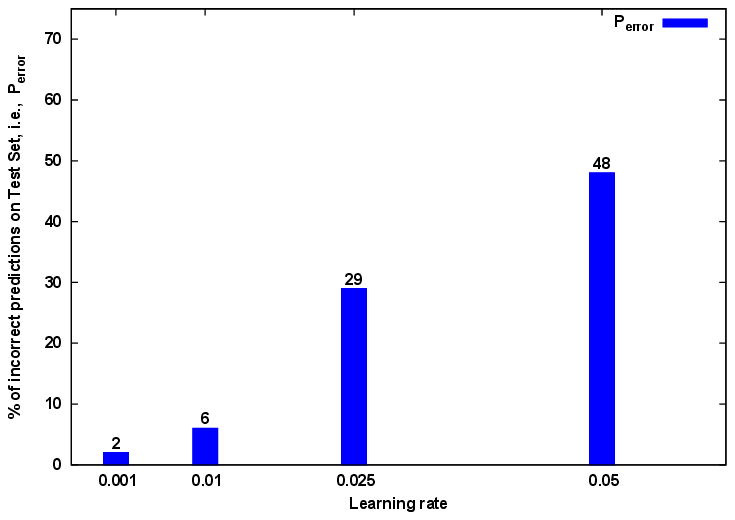}
\caption{Impact of increase in learning rate}
\label{fig_increase_learning_rate_inkscape}
\end{figure}

As seen in Fig. \ref{fig_increase_learning_rate_inkscape} increase in learning rate leads to major degradation in failure prediction by the classifier.
\subsection{Failure prioritization}
\begin{figure}[ht]
\centering
\includegraphics[width=\columnwidth]{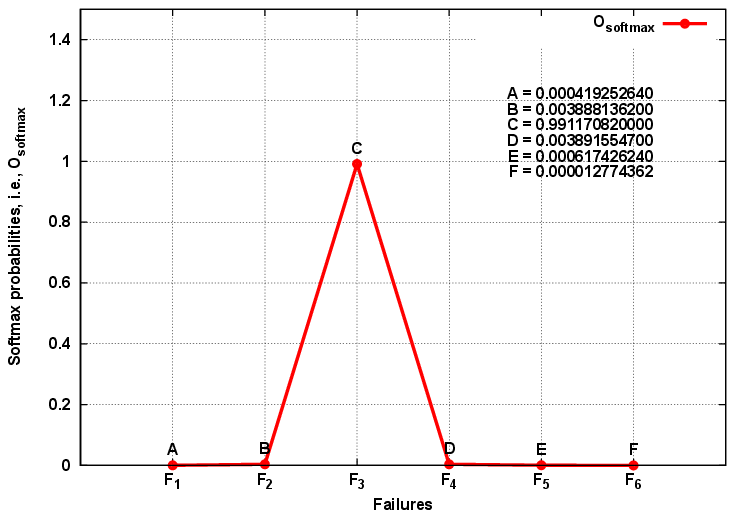}
\caption{Softmax probabilities of failures, i.e, $O_{softmax}$}
\label{fig_ahp_dnn_prob_before_shape_inkscape}
\end{figure}

\begin{figure}[ht]
\centering
\includegraphics[width=\columnwidth]{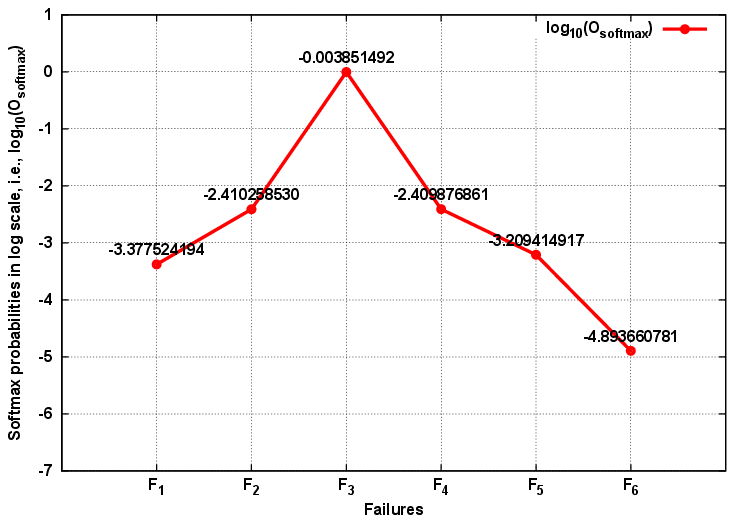}
\caption{Softmax probabilities of failures in log scale, i.e, $log_{10}(O_{softmax})$}
\label{fig_ahp_dnn_prob_before_shape_log_inkscape}
\end{figure}

\begin{figure}[ht]
\centering
\includegraphics[width=\columnwidth]{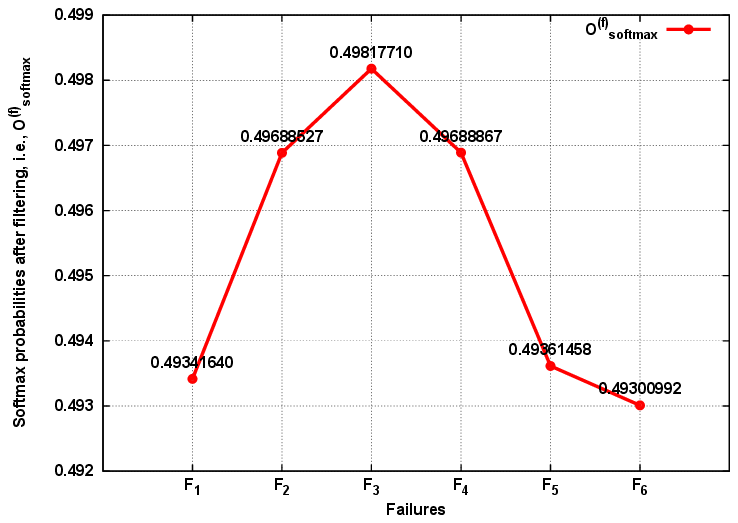}
\caption{Softmax probabilities of failures after filtering, i.e., $O^{(f)}_{softmax}$}
\label{fig_ahp_dnn_prob_after_shape_inkscape}
\end{figure}

To explain the functionality of failure prioritization concisely, the event failure matrix $I_{6,10}$ (\ref{eqn_ahp_input_event_failure_matrix}) is considered which has 6 failures along rows and 10 events as columns. However, the prioritization mechanism will work for a higher number of failures and events as well. Matrix $I_{6,10}$ is the input to the multi-class classifier and the corresponding output is matrix $O_{6,6}$ (\ref{eqn_ahp_output_failures}). These two matrices are used for training the classifier applying GA, mapping, and normalization steps (as explained above). Each row of  $I_{6,10}$ corresponds to the same row in $O_{6,6}$.
\begin{equation}
I_{6,10} = \kbordermatrix{
        & E_1 & E_2 & E_3 & E_4 & E_5 & E_6 & E_7 & E_8 & E_9 & E_{10} \\
    F_1 & 1 & 1 & 0 & 0 & 1 & 1 & 1 & 1 & 1 & 0\\
    F_2 & 0 & 0 & 1 & 1 & 1 & 1 & 1 & 0 & 0 & 0\\
    F_3 & 1 & 1 & 1 & 1 & 1 & 1 & 0 & 1 & 1 & 0\\
    F_4 & 0 & 1 & 0 & 0 & 1 & 1 & 1 & 0 & 1 & 0\\
    F_5 & 0 & 1 & 1 & 0 & 1 & 1 & 1 & 1 & 0 & 0\\
    F_6 & 1 & 1 & 0 & 0 & 0 & 0 & 1 & 1 & 1 & 0
  }
  \label{eqn_ahp_input_event_failure_matrix}
\end{equation}
\begin{equation}
O_{6,6} = \kbordermatrix{
    \\
    & 1 & 0 & 0 & 0 & 0 & 0\\
    & 0 & 1 & 0 & 0 & 0 & 0\\
    & 0 & 0 & 1 & 0 & 0 & 0\\
    & 0 & 0 & 0 & 1 & 0 & 0\\
    & 0 & 0 & 0 & 0 & 1 & 0\\
    & 0 & 0 & 0 & 0 & 0 & 1
  }
  \label{eqn_ahp_output_failures}
\end{equation}
After training, when the vector $I_{vector}$ in (\ref{eqn_ahp_sample_input}) is presented as input to $L_{input}$, it predicts a failure with softmax output probabilities ($p_{F_1}, p_{F_2},..,p_{F_6}$) in (\ref{eqn_ahp_sample_output_softmax}) which has highest value (underlined) for third entry and lower values for others. With decision threshold $D_{thres}$ = 0.5, (\ref{eqn_ahp_sample_output_softmax}) is transformed to (\ref{eqn_ahp_sample_output}) which is same as third row in (\ref{eqn_ahp_output_failures}) and corresponds to the correct prediction of $F_3$ in (\ref{eqn_ahp_input_event_failure_matrix}). The probabilities of (\ref{eqn_ahp_sample_output_softmax}) are plotted in Fig. \ref{fig_ahp_dnn_prob_before_shape_inkscape}. Since, one but the rest of the values are very low the same is plotted in logarithmic scale in Fig. \ref{fig_ahp_dnn_prob_before_shape_log_inkscape} to understand the variations.
\begin{equation}
I_{vector} = \kbordermatrix{
    \\
    & 1 & 1 & 1 & 1 & 1 & 1 & 0 & 1 & 1 & 0
    }
    \label{eqn_ahp_sample_input}
\end{equation}
\begin{equation}
O_{softmax} = \kbordermatrix{
    \\
    p_{F_1} & 4.1925264 \times 10^{-4}\\
    p_{F_2} & 3.8881362 \times 10^{-3}\\
    p_{F_3} & \underline{9.9117082 \times 10^{-1}}\\
    p_{F_4} & 3.8915547 \times 10^{-3}\\
    p_{F_5} & 6.1742624 \times 10^{-4}\\
    p_{F_6} & 1.2774362 \times 10^{-5}
    }
    \label{eqn_ahp_sample_output_softmax}
\end{equation}
\begin{equation}
O_{vector} = \kbordermatrix{
    \\
    & 0 & 0 & 1 & 0 & 0 & 0
    }
    \label{eqn_ahp_sample_output}
\end{equation}
As explained above, it is necessary to apply the shape-preserving filter on the output softmax probabilities (\ref{eqn_ahp_sample_output_softmax}) to reduce the variance. The output (\ref{eqn_ahp_sample_output_softmax_filter}) of the shape preserving filter applied on (\ref{eqn_ahp_sample_output_softmax}) with $\delta = 0.001$ is shown in Fig. \ref{fig_ahp_dnn_prob_after_shape_inkscape}. Note that after filtering $p^{(f)}_{F_3} > p^{(f)}_{F_4} > p^{(f)}_{F_2} > p^{(f)}_{F_5} > p^{(f)}_{F_1} >  p^{(f)}_{F_6}$  in (\ref{eqn_ahp_sample_output_softmax_filter}) which is in the same order as $p_{F_3} > p_{F_4} > p_{F_2} > p_{F_5} > p_{F_1} > p_{F_6}$ in (\ref{eqn_ahp_sample_output_softmax}) as expected. Only the variance is reduced after filtering. Hence, the predicted failure remains the same after filtering which is $F_3$.
\begin{equation}
O^{(f)}_{softmax} = \kbordermatrix{
    \\
    p^{(f)}_{F_1} & 0.49341640\\
    p^{(f)}_{F_2} & 0.49688527\\
    p^{(f)}_{F_3} & \underline{0.49817710}\\
    p^{(f)}_{F_4} & 0.49688867\\
    p^{(f)}_{F_5} & 0.49361458\\
    p^{(f)}_{F_6} & 0.49300992
    }
    \label{eqn_ahp_sample_output_softmax_filter}
\end{equation}

\begin{figure}[ht]
\centering
\includegraphics[width=\columnwidth]{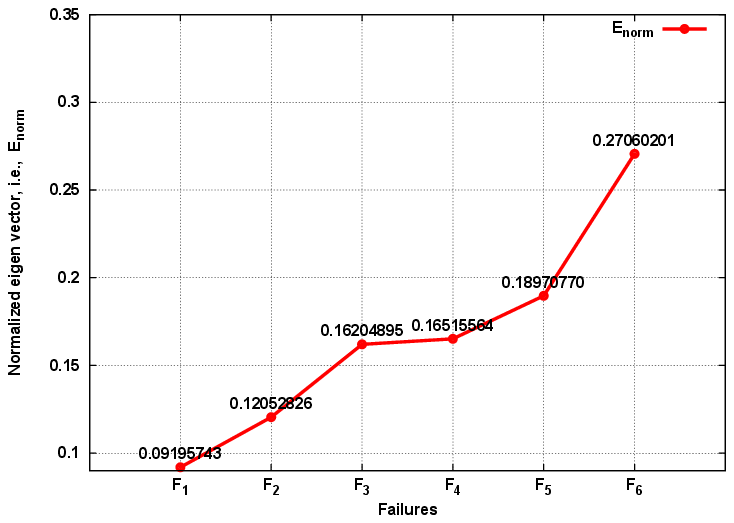}
\caption{Normalized eigen vector, i.e., $E_{norm}$}
\label{fig_eigen_vector_before_shape_inkscape}
\end{figure}

\begin{figure}[ht]
\centering
\includegraphics[width=\columnwidth]{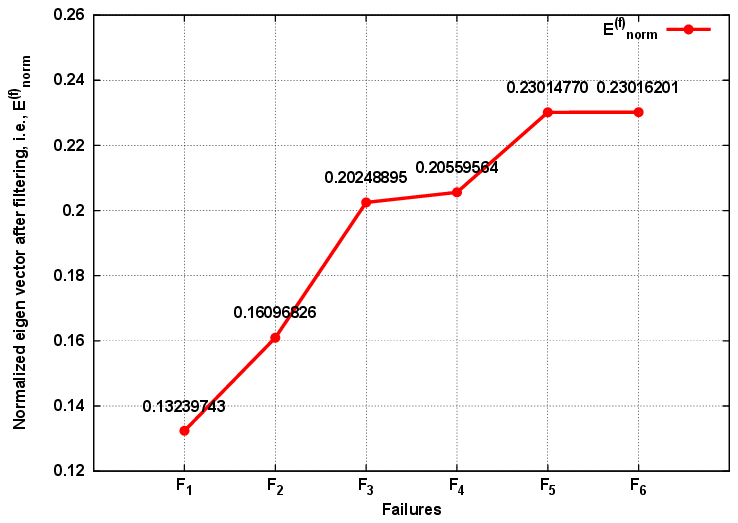}
\caption{Normalized eigen vector after filtering, i.e., $E^{(f)}_{norm}$}
\label{fig_eigen_vector_after_shape_inkscape}
\end{figure}

\begin{figure}[ht]
\centering
\includegraphics[width=\columnwidth]{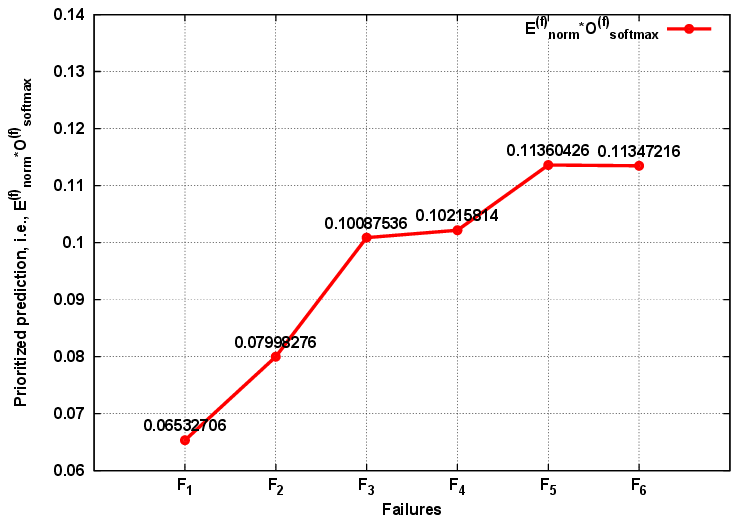}
\caption{Prioritized prediction, i.e., $E^{(f)}_{norm}$ * $O^{(f)}_{softmax}$}
\label{fig_ahp_adj_weights_adj_probs_inkscape}
\end{figure}

Based on human judgment (e.g., business needs), the pairwise importance matrix of the 6 failures is provided in (\ref{eqn_ahp_sample_pairwise}).
\begin{equation}
C_{6,1} = \kbordermatrix{
    \\
    & 1 & 1/2 & 3/5 & 5/6 & 3/7 & 3/8\\
    & 2/1 & 1 & 2/3 & 4/5 & 3/4 & 2/9\\
    & 5/3 & 3/2 & 1 & 5/6 & 10/11 & 11/15\\
    & 6/5 & 5/4 & 6/5 & 1 & 13/15 & 15/17\\
    & 7/3 & 4/3 & 11/10 & 15/13 & 1 & 17/19\\
    & 8/3 & 9/2 & 15/11 &  17/15 & 19/17 & 1\\
  }
  \label{eqn_ahp_sample_pairwise}
\end{equation}
Normalized eigen vector obtained using the power method \cite{cite_eigen_power_method} is shown in (\ref{eqn_ahp_norm_eigen_vector}). The values in the vector (\ref{eqn_ahp_norm_eigen_vector}) act as weights, $w_{F_1}, w_{F_2},.., w_{F_6}$, for the corresponding failures. The weights are plotted in Fig. \ref{fig_eigen_vector_before_shape_inkscape}. The highest value underlined in (\ref{eqn_ahp_norm_eigen_vector}) is the weight $w_{F_6}$ for failure $F_6$ which has the maximum priority according to the business needs. The weights being in ascending order is mere coincidence and not intentional.
\begin{equation}
E_{norm} = \kbordermatrix{
    \\
    w_{F_1} & 0.09195743\\
    w_{F_2} & 0.12052826\\
    w_{F_3} & 0.16204895\\
    w_{F_4} & 0.16515564\\
    w_{F_5} & 0.18970770\\
    w_{F_6} & \underline{0.27060201}
    }
    \label{eqn_ahp_norm_eigen_vector}
\end{equation}
The eigen vector is passed through the shape-preserving filter as done above for the probabilities. The filtered values are shown in (\ref{eqn_ahp_norm_eigen_vector_filter}) and plotted in Fig. \ref{fig_eigen_vector_after_shape_inkscape}. The filtering reduces the variance in values and maintains the same relative ordering as expected.
\begin{equation}
E^{(f)}_{norm} = \kbordermatrix{
    \\
    w^{(f)}_{F_1} & 0.13239743\\
    w^{(f)}_{F_2} & 0.16096826\\
    w^{(f)}_{F_3} & 0.20248895\\
    w^{(f)}_{F_4} & 0.20559564\\
    w^{(f)}_{F_5} & 0.23014770\\
    w^{(f)}_{F_6} & \underline{0.23016201}
    }
    \label{eqn_ahp_norm_eigen_vector_filter}
\end{equation}
The vectors (\ref{eqn_ahp_sample_output_softmax_filter}) and (\ref{eqn_ahp_norm_eigen_vector_filter}) are multiplied elementwise and shown in (\ref{eqn_ahp_norm_eigen_vector_prob_filter_product}) (also plotted in Fig. \ref{fig_ahp_adj_weights_adj_probs_inkscape}) with the highest value underlined. The highest value is the predicted failure which in this case is $F_5$. Thus, the original prediction of $F_3$ by the classifier changes to $F_5$ based on business needs.
\begin{equation}
E^{(f)}_{norm}*O^{(f)}_{softmax} = \kbordermatrix{
    \\
    & 0.06532706\\
    & 0.07998276\\
    & 0.10087536\\
    & 0.10215814\\
    & \underline{0.11360426}\\
    & 0.11347216
    }
    \label{eqn_ahp_norm_eigen_vector_prob_filter_product}
\end{equation}
\subsection{Timing measurements}
For the configuration in Table \ref{table_results_hidden_layers} the training time is around 12 minutes 30 seconds on a 12 logical core 64 GB RAM machine without GPU. Generation of the artificial data set (for training and testing the classifier) of size 100,000 takes around 1 hour and 10 minutes.  The shape-preserved filtering over 1000 softmax failure probabilities takes around 164 milliseconds. This filter timing is on the higher side and will investigated further in the future.

\subsection{Cross entropy loss}
\begin{figure}[ht]
\centering
\includegraphics[width=\columnwidth]{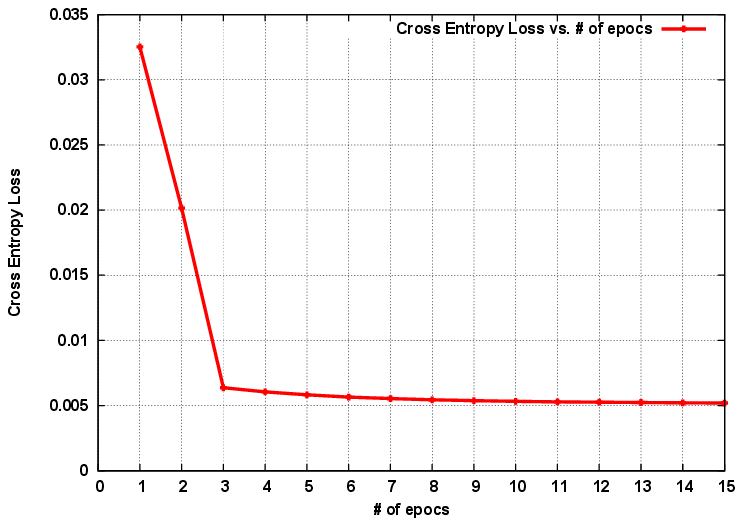}
\caption{Cross Entropy Loss}
\label{fig_cross_entropy_loss_inkscape}
\end{figure}

Fig. \ref{fig_cross_entropy_loss_inkscape} shows the cross entropy loss (along \emph{y}-axis) versus epochs (along \emph{x}-axis) which indicates a decreasing trend during training of the classifier as expected. Gradient descent is performed by applying the \verb|adam| algorithm to minimize the loss function. The parameters for \verb|adam| are provided in Table \ref{table_adam_parameters}.
\begin{table}[ht]
  \caption{Parameters for Adam algorithm}
  \centering
  \begin{tabular}{|p{2cm}|p{3cm}|}
  \hline
  Parameter & Values \\  [0.5ex]
  \hline
  learning rate & 0.001 \\
  \hline
  $\beta_1$ & 0.900 \\
  \hline
  $\beta_2$ & 0.999\\
  \hline
  $\epsilon$ & $10^{-7}$\\
  \hline
  \end{tabular}
  \label{table_adam_parameters}
\end{table}

\section{Conclusion}\label{section_conclusion}
This paper presented an NN-based multi-class classifier for non-intrusive failure prediction for systems that are deployed in the field using artificially generated data sets and keeping the actual data entirely private. Key texts and their sequences in the private system logs which lead to failures are encoded as binary bits (events). The data set to train and test the classifier is artificially generated using these binary events applying pattern repetition, steps from GA, and random sampling from disjoint sets of positive real numbers. To prioritize the failures based on business needs, AHP is used to define their weights and then a shape-preserving filter is applied to both the weight and the softmax output vectors. The \emph{argmax} of the elementwise product of two vectors is the final prioritized predicted failure. Results reveal that the failure prediction works with very high accuracy. On a broader scope, any classification problem is solvable where input features can be mapped to binary values and have one-hot vectors as output using the proposed mechanism with the artificially generated data set, keeping the actual data private to the product/data owners and providing classification-as-a-service.

Future work will explore other types of neural networks, activation functions, training mechanisms, GA variants, MCDM methods, and shape-preserving filters.

\section*{Declarations}
\begin{itemize}
\item \emph{Funding:} This project is sponsored by Tejas Networks Ltd., Bangalore, India
\item \emph{Conflict of interest/Competing interests:} The authors have no relevant financial or non-financial interests to disclose.
\item \emph{Ethics approval and consent to participate:} No living beings has been used in this research.
\item \emph{Consent for publication:} All the authors have consented for this submission and possible publication.
\item \emph{Data availability:} No data is made available.
\item \emph{Materials availability:} No material is applicable.
\item \emph{Code availability:} No code is made available.
\item \emph{Author contribution:} Dibakar Das is responsible for conceptualization of the idea, design, implementation, analysis of results and preparation of this manuscript. Vikram Seshasai, Vineet Sudhir Bhat and Pushkal Juneja from Tejas Networks Ltd. provided valuable inputs to evaluate the mechanism and reviewing the work. Jyotsna Bapat and Debabrata Das were responsible for managing the project and reviewing the manuscript.
\end{itemize}

\bibliography{net_failure_pred}

\end{document}